\documentclass[11pt]{article}
\usepackage[T1]{fontenc}
\usepackage{consilr2024}
\usepackage{times}
\usepackage{url}
\usepackage{latexsym}
\usepackage{authblk}
\usepackage{hyperref}
\usepackage{etoolbox}
\usepackage{graphicx}
\usepackage{todonotes}
\usepackage{float}
\usepackage{subcaption}

\newcommand{\keyword}[1]{
\hspace{0.2cm}%
\fontsize{10}{12}\selectfont%
\textbf{Keywords: } %
}

\title{\textbf{Exploring Large Language Models for Translating Romanian Computational Problems into English}}

\author[1]{Adrian Marius Dumitran}
\author[2]{Adrian-Catalin Badea}
\author[3]{Stefan-Gabriel Muscalu}
\author[4]{Angela-Liliana Dumitran}
\author[5]{Stefan-Cosmin Dascalescu}
\author[6]{Radu-Sebastian Amarie}

\affil[1]{University of Bucharest, Softbinator
\break
\texttt{marius.dumitran@unibuc.ro}}

\affil[2]{University of Bucharest, UiPath
\break
\texttt{badeaadi1999@gmail.com}}

\affil[3]{It Just Works Inc.
\break
\texttt{stefan.gabriel.muscalu@gmail.com}}

\affil[4]{University of Bucharest
\break
\texttt{dumitranangela@gmail.com}}

\affil[5]{QPillars, University of Bucharest
\break
\texttt{stefdasca@gmail.com}}

\affil[6]{It Just Works Inc., University of Bucharest
\break
\texttt{raduamarie@gmail.com}}

\date{}

\begin{document}
\maketitle
\begin{abstract} 

Recent studies have suggested that large language models (LLMs) underperform on mathematical and
computer science tasks when these problems are translated from Romanian into English, compared to
their original Romanian format. Accurate translation is critical for applications ranging from automatic
translations in programming competitions to the creation of high-quality educational materials, as well as
minimizing errors or fraud in human translations.
This study shows that robust large language models (LLMs) can maintain or even enhance their performance in translating less common languages when given well-structured prompts. Our findings suggest that LLMs, with appropriate supervision, can be reliably used for the automatic translation of
IOI (International Olympiad in Informatics)-style tasks. We evaluate several translation methods across
multiple LLMs, including OpenRoLLM, Llama 3.1 8B, Llama 3.2 3B and GPT-4o, assessing their
translation accuracy and performance stability through repeated runs. Additionally, we augment the
OJI (Romanian County-Level Informatics Olympiad) Romanian dataset with accurate English translations,
enhancing its utility for future LLM training and evaluation.
Through detailed syntactic and semantic analyses, we confirm that with human oversight, LLMs can serve
as a viable solution for multilingual problem-solving. We also compare the translation quality of LLMs
against human translators, as evaluated by a certified expert, underscoring the potential of LLMs in realworld scenarios.

\end{abstract}


\section{Introduction}

Large language models (LLMs) have showcased remarkable capabilities across a wide range of natural language processing (NLP) tasks, including text generation, translation, code completion, and problem-solving in technical domains. Despite this progress, recent research has pointed to challenges in applying LLMs to structured domains, particularly in areas such as mathematics and computer science. Studies by \cite{rae2021} indicate that while scaling LLMs improves performance across many tasks, structured tasks like mathematical problem-solving see smaller gains, suggesting that complexities in these domains are not fully captured or preserved. This discrepancy prompts further exploration of linguistic and computational factors that impact performance when translating such structured tasks across languages.

Translating mathematical and computational tasks across languages presents challenges that go beyond typical linguistic differences. Technical problems require high levels of precision, and even minor translation errors—whether due to loss of mathematical context or subtle language ambiguities—can prevent humans from correctly understanding and solving these problems. When translations are flawed, they can hinder human problem-solvers by introducing confusion or misinterpretation, which is particularly impactful in high-stakes, multilingual environments.


In \cite{RoMath:24} emphasize the need for dedicated resources beyond simple automatic translation, particularly for underrepresented languages like Romanian. Other researches such as \cite{RoCode:24} argue for the necessity of developing code models for languages other than English, highlighting the current limitations of large language models in understanding and solving problems in non-English languages. 


In \cite{OJI:24}, we delved into the performance of English large language models (LLMs) in solving competitive programming problems from the Romanian Informatics Olympiad at the county level. The study revealed significant variations in LLM performance across different grades and problem types, with GPT-4 showing strong performance.

The primary objective of this paper is to systematically analyze how translation-induced errors affect the ability of humans to solve technical problems when using LLMs as a translation aid. Focusing on Romanian-to-English translations of IOI-style problems from the OJI  dataset, we evaluate different translation strategies to identify the most reliable methods. Through repeated testing and performance analysis, we determine the best ways to ensure that translations retain the original meaning and clarity. In order to support human problem-solvers and to help LLM training, we enhance the OJI dataset with accurate English translations and propose an optimized prompt that improves translation accuracy, ensuring that humans can effectively solve the translated problems with the same or better understanding than from the original Romanian versions.

Although prior research exists on Romanian-English translation and the translation of mathematical problems, we have not identified any studies specifically addressing Machine Translation of computer science tasks between any language pairs. Furthermore, we are confident that no such work has been conducted for the Romanian-English language pair.

This paper is organized as follows: Section 2 outlines the methodology, detailing the translation approaches and the LLM evaluation process. Section 3 presents the results of our experiments, with a focus on translation accuracy and performance variability. In Section 4, we conduct a syntactic and semantic analysis of translation errors, and Section 5 introduces the optimal prompt and discusses the translation improvements it brings. Finally, Section 6 presents the conclusions.

\section{Methodology}

The methodology for this study was structured into several key stages, employing both quantitative and qualitative methods to thoroughly analyze the data. 

\begin{enumerate}


\item In the initial stage, we chose 44 problem statements in Romanian from the same grade (15-16 years old), out of the 300 in \href{https://huggingface.co/datasets/OpenLLM-Ro/Ro-OJI}{"OJI" dataset} introduced by ~\cite{OJI:24}. 

\item In the second stage, we computed a \textbf{"Ro\_score"} for each problem, representing the highest score GPT-4o achieved after attempting to solve each problem five times in its original Romanian form.

\item The third stage involved translating the problems using a diverse set of LLMs(which will be detailed in \ref{ssec:llm_selection} , with variations in temperature settings for some models to assess their impact on translation quality.

\item In the fourth stage, GPT-4o was run five times with temperature 0.4 on each translated version of the problem, and scores were obtained for each task and each translation.

\end{enumerate}
Importantly, we compared the scores achieved by GPT-4o on the original Romanian tasks (Ro\_score) and the translated tasks, with GPT-4o acting as the evaluator while the other LLMs handled only the translation process, without generating code.

A quantitative analysis was conducted to compare the judge scores across all LLM translations, with additional investigations into the effects of temperature settings and model size.

Next, we conducted an error analysis, where translations were reviewed and categorized by a linguist. Common issues identified included mistranslations, inappropriate language, untranslated content, inconsistent translation of algorithmic and computational terminology, and untranslated examples. Additionally, members of the OJI scientific committee examined the translations for technical issues related to the content.

Finally, we conducted a comparison between human translations provided by members of the OJI scientific committee and those generated by LLMs, to evaluate how closely the LLM translations align with expert human translations.

\subsection{LLM Selection}\label{ssec:llm_selection}

For our experiments, we selected multiple state-of-the-art LLMs that are widely used for natural language processing and problem-solving tasks:
\begin{itemize}
   \item \textbf{Llama 3.1 8B}: A lightweight LLM optimized for efficiency and performance \cite{dubey2024llama3herdmodels}.
   \item \textbf{Llama 3.2 3B}: A really lightweight LLM optimized for efficiency and performance \cite{dubey2024llama3herdmodels}.
   \item \textbf{GPT-4o}: The versatile flagship from Azure OpenAI, highly capable, known for its broad generalization abilities \cite{ChatGPT}.
   \item \textbf{OpenLLMRo}: We tried multiple models from the \href{https://huggingface.co/collections/OpenLLM-Ro}{OpenLLMRo community} \cite{OpenLLMRo2024}.

   \item \textbf{panSophic-1 preview}: A Romanian language-specific \href{https://pansophic.ai/} {model}.
   \item \textbf{Aya35B}: Aya 35B, released by Cohere, is part of a new family of state-of-the-art multilingual models. One of the 23 languages that it supports is Romanian \cite{Aya23}.
\end{itemize}
   We conducted additional tests on Mistral 7B, Gemma 7B, and Gemini 1.5Pro, but the results for these models were below average.
   
Most of our models fit into memory and no quantization was needed except for the Aya35B model. According to \cite{Marchisio2023}, Aya35B shows minimal degradation in translation tasks (only -0.7\% on the Flores benchmark). 

GPT-4o was chosen for being state of the art, while the open-weight models were selected for their perceived strong performance in translating or knowledge of Romanian.

\subsection{Translation process}\label{ssec:translation_process}

We translated each task using the LLMs listed in the previous section. 

 \cite{Peeperkorn2024} investigate the impact of temperature adjustments on the creativity of outputs produced by large language models. The study demonstrates that setting the temperature above 1 has minimal effect on the results, specifically in the context of story generation.

For each task and model, we initially performed one translation at a temperature setting of 0.6. For the top-performing models, we repeated the translations at varying temperature settings: 0.2, 0.6, and 1.0 to assess performance across different levels of randomness and across multiple iterations. 

We started with a temperature of 0.6 as a balanced approach, providing a mix of randomness and certainty in the model's predictions. This allowed us to get a general sense of the model's performance.

The top-performing models were then tested at different temperature settings to further understand their capabilities. A temperature of 0.2 was chosen to see how the model performs when it is more confident in its predictions. This could potentially lead to more accurate translations, but it could also result in less creative or diverse outputs.

On the other hand, a temperature of 1.0 was chosen to push the model towards more randomness in its predictions. This could lead to more diverse and creative translations, but it could also result in less accuracy.

\subsubsection{Prompt}

The following simple prompt was used for translation with all models: \newline

\textit{You will be provided with an OJI (Olimpiada Județeană de Informatică) challenge and you will need to translate it from Romanian to English.\\
Rules: Any strings related to the body of the challenge should not be translated, such as input/output examples or file names.\\
The translation should follow the original format (numbering of paragraphs, examples, etc).\\
The translation should be in the same tense as the original text.\\
Respond only with the challenge body in English.}

This prompt was chosen for its specific requirements, including the need to preserve the structure, strings, and tense of the original text. These constraints make the task an interesting subject for studying translation strategies and the challenges of translating technical and domain-specific content.

We aimed for a simple, reasonable prompt that typical LLM users could employ. However, as shown later, a more complex prompt may yield better results.

Moreover, the task's focus on the OJI challenges also provides an opportunity to explore the translation of educational and competitive programming content, which is a less-studied area in translation studies.

\subsection{Problem Selection}

Most of the tests have been done on problems of low to medium difficulty, as our main method of automatic verification was to have the translated tasks solved by LLMs. We focus our analysis on 8th grade problems, which can be found on \cite{kilonova}. Kilonova is a renowned Romanian online judge platform, specifically designed for training in computer science olympiads. It has gained significant popularity in recent years due to its comprehensive collection of problems from all past olympiads. The platform provides an interactive and challenging environment for students to enhance their problem-solving and programming skills.

8th grade problems tend to have complex texts with many subproblems. The 8th grade curriculum also includes string-related problems, and we know that such problems tend to cause difficulties in translation due to commonly mistranslated terms, such as: "subsir" (subsequence, not necessarily a contigous part of the sequence) and "subsecvență" (subarray or substring). 

We offer translations for all problems in the OJI dataset~\cite{OJI:24} and provide an external link to a broader translation dataset.

\section{Translation Comparison}

\begin{figure}[htbp]
    \centering
    \includegraphics[width=1\textwidth]{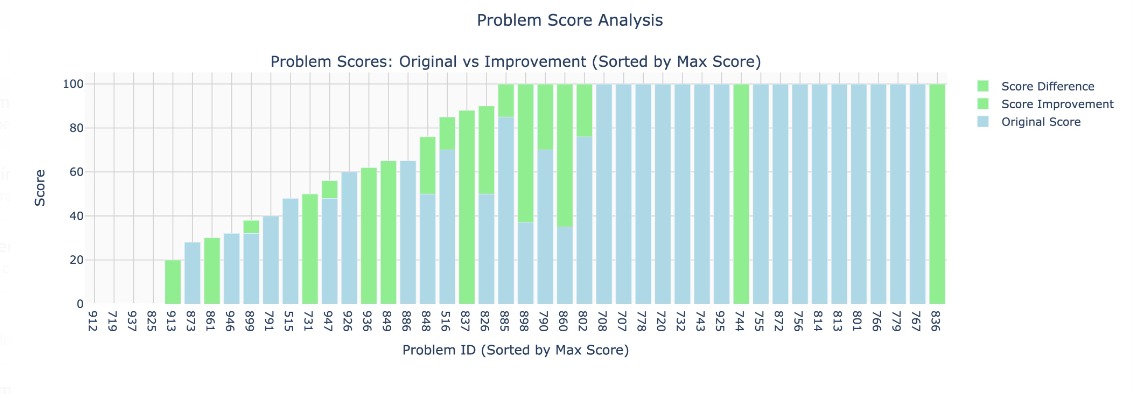}
    \caption{Comparison of maximum translation scores versus the original score (ro\_score) across all problems.}
    \label{fig:all_vs_ro_temp}
\end{figure}

We start by emphasizing that, to our surprise, translating from Romanian to English does not seem to cause performance issues. Moreover, there may even be benefits to translating the task into English.
\newpage
Figure \ref{fig:all_vs_ro_temp}
 illustrates the difference between the maximum translation score across all models and the ro\_score, showing this difference for each problem individually. For 22 out of 38 problems, the ro\_score matches the maximum translation score, while for the remaining cases, the maximum translation score exceeds the initial score.

\subsection{Automated verification}
 Our methodology involved conducting five runs using GPT-4o to solve the initial Romanian tasks. After obtaining the baseline scores, the problem statements were translated into English using various LLMs and temperatures. Although increasing the number of runs improves the likelihood that an LLM will solve a given problem, it also incurs additional costs. Based on our findings in \cite{OJI:24}, we conclude that five runs provide a good balance between quality and cost. For lower temperatures, fewer runs are also effective.

The translated versions were then fed back into GPT-4o, where five additional runs were performed for each translated text. We compared the results of these runs with the initial Romanian versions. If a similar or better score was achieved on the translated English version, we considered the translation to be effective.

In some cases, better scores were observed for the English translations due to several factors, including:
    
    \begin{itemize}
        \item \textbf{Randomness in the outputs}: We often used temperatures higher than 0, introducing randomness that can sometimes lead to better outcomes in the English version.
        \item \textbf{Superior training data in English}: The model is typically trained on larger and more diverse English datasets, leading to better performance in English tasks.
        \item \textbf{Refined error handling in English}: LLMs generally have more robust error detection and correction mechanisms when processing English inputs, improving their overall problem-solving capabilities.
        \item \textbf{Better understanding of formal structures in English}: English-specific models are better at understanding and interpreting the formal language structures often used in technical problems.
    \end{itemize}

\subsubsection{Overall results}

\begin{figure}[H]
    \centering
    \includegraphics[width=1.00\linewidth]{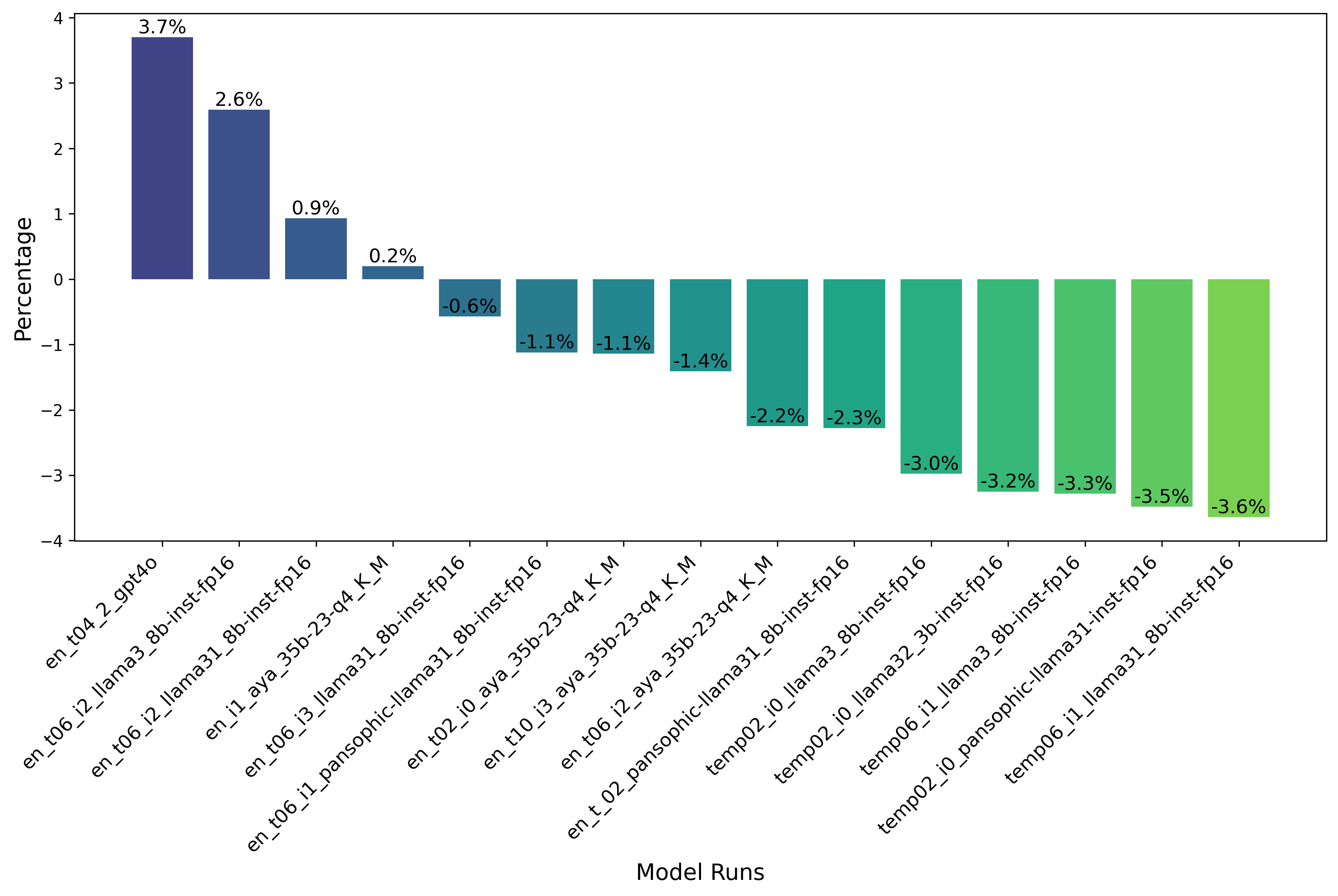}
    \caption{Top 15 model runs vs RoScore}
    \label{fig:top_runs}
\end{figure}

Figure \ref{fig:top_runs} presents a comparative analysis of the 15 highest-scoring model runs, arranged in descending order of performance. The x-axis identifies each model run with an alphanumeric code, which represents specific model architectures, temperature settings \textit{('t')}, iteration number \textit{('i')}, and, where known, the number of parameters. The percentages above the bars show the performance difference compared to the Ro\_score.

The close clustering of scores, ranging from +3.7\% to -3.6\%, illustrates that top models achieved similar performance levels, with only marginal variations between them. Unsurprisingly, multiple runs with the same settings can yield different results due to the inherent randomness in model outputs. OpenLLmRo models were excluded from this analysis as their focus on Romanian significantly impacted their English translation capabilities.

These results underscore the diminishing returns at the upper end of model performance and suggest that further progress in enhancing translation accuracy may depend on fine-tuning models for specific tasks or further improving prompt design, rather than purely increasing model size or complexity.

\subsection{Temperatures - the creativity parameter}

\begin{table}[h]
\centering
\begin{tabular}{|c|c|c|c|c|}
\hline
Translation Model & Temperature & Size & Scores & Score Average \\
\hline
gpt4o & 0.6 & * & 58.84 & 58.84\\
ro\_score & 0.6 & * & 55.13 & 55.13\\
aya & 0.2, 0.6, 1.0 & 35B & 54.00, \textbf{55.34}, 53.73 & 54.35 \\
llama3 & 0.2, 0.6, 1.0 & 8B & 52.16, \textbf{57.73}, 47.45 & 52.44\\ 
llama31 & 0.2, 0.6, 1.0 & 8B & 50.05, \textbf{56.07}, 49.77 & 51.96 \\
llama32 & 0.2, 0.6, 1.0  & 3B & \textbf{51.89}, 51.41, 49.41 & 50.90 \\
pans-llama31 & 0.2, 0.6, 1.0 & 8B & 52.86, \textbf{54.02}, 46.93 & 51.27 \\
gemini & 0.6 & * & 45.11 & 45.11 \\
mistral & 0.6 & 7B & 44.50 & 44.50 \\
gemma & 0.6 & 7B & 43.77 & 43.77\\
\hline
\end{tabular}
\caption{Average result on dataset over temperature and size}
\label{tab:model_performance}
\end{table}

For the models for which we ran more iteratios, the scores are aligned respectively to the temperature in the table \ref{tab:model_performance}

We chose temperatures of 0.2, 0.6, and 1.0, as discussed in \ref{ssec:translation_process}. 
We only ran temperature 0.2 and 1.0 for the LLMs with good results on temperature 0.6. Pricier models such as gpt4o or gemini were run only once, with 0.6. Key findings: 

\begin{enumerate} 

\item \textbf{(Optimal temperature)}: On average temperature 0.6 showed best results followed by 0.2 and 1.0.
\item  \textbf{(Small Model)}:
     Llama 3.2 3B, our smallest model, was the only one that performed better at a temperature of 0.2, indicating that smaller models tend to deviate from optimal results more quickly.
\item \textbf{(Temperature 0.6 similar text)}. We add two translations by Llama 3.1 at temperature 0.6:\newline
    \textit{A word consisting only of small letters is given. We call anagram a word formed by rearranging the letters of the given word. For example, armata is an anagram of tamara. Obviously, a word can be considered its own anagram.} \newline
    \textit{A word consisting only of lowercase letters is given. We call anagram a word formed by the letters of the given word, rearranging them if necessary. For example, armata is an anagram of tamara. Obviously, a word can be considered an anagram of itself.}

    Note that while the story is a bit modified in translation the rest of the text(not included here) is almost identical (the part related to the actual task and the restrictions). Thus, models tend to produce similar but slightly varied translations at temperature 0.6.
\end{enumerate}

\subsection{Model size} 

We utilized language models of varying sizes to represent different capacities and computational requirements (see \ref{tab:model_performance}
). The models, ordered by size, include Llama 3.2 3B, Gemma 7B, Mistral 7B, Aya 8B, Llama 3 8B, Llama 3.1 8B, Pansophic (based on Llama 3.1 8B), Aya 35B, and GPT-4o (estimated at over 100 billion parameters). By incorporating models ranging from 3 billion to over 100 billion parameters, we investigated how model size impacts translation performance across temperature settings. This range allowed us to examine the trade-offs between computational resources and translation quality, as well as how different-sized models respond to temperature adjustments in language translation tasks.

\textbf{Key Observations}

\begin{enumerate}
    
  \item \textbf{GPT-4o at 100B+}:
    \begin{itemize}
        \item GPT-4o, with over 100 billion parameters, demonstrates the highest performance in our study. However, GPT-4o's scores should be evaluated in comparison with other large models like Llama 3.1 405B. Additionally, GPT-4o's performance may be influenced by the fact that it was both the translation model and the judge, potentially giving it an advantage by aligning the translation with the evaluation model’s own language patterns.
    \end{itemize}

\item \textbf{Smaller Models Struggle with Longer Tasks}
    \begin{itemize}
        \item In our analysis, smaller models often struggled to complete translations, especially for longer, more complex tasks. Many of the tasks that received a score of 0 points after translation were incomplete. This was particularly evident in problems with a lot of restrictions and examples like \href{https://kilonova.ro/problems/515}{problem 515}, where examples and restrictions were either omitted or poorly translated. For instance, in one Aya translation, a big part of the text was omitted and instead the following text offered: \textit{The rest of the challenge remains unchanged and can be found in your original text}, instead of translating the task and restrictions fully.
    \end{itemize}

\item \textbf{Llama Models' Performance at 8B and 3B}
    \begin{itemize}
        \item The Llama models, known for robust multilingual capabilities, consistently succeeded in translation tasks, often matching or exceeding the original Romanian scores (Ro\_score). Zero scores were only due to incomplete translations. This suggests that with better prompts tailored to smaller models and allowing multiple attempts, both Llama 3 and Llama 3.1 could potentially match GPT-4o's performance.
        
        \item However, Llama 3.2 3B underperformed in comparison to the other Llama models, proving inadequate for this specific task.
    \end{itemize}

\item \textbf{Similar Translation Quality Across Top Models}:
    \begin{itemize}
        \item As detailed in \ref{sec:grammatical_analysis}, our qualitative analysis revealed no significant differences in translation quality among the top-performing models. The variations in performance were largely attributed to model fatigue when handling longer tasks.
    \end{itemize}

\end{enumerate}

\section{Grammatical Analysis} \label{sec:grammatical_analysis}
Various translation techniques, including expansion, adaptation, transposition, and structure shift, continue to be employed by machine translation (MT) systems. As noted by \cite{Wu2016}, these systems utilize large datasets and deep learning algorithms to effectively implement these techniques, aiming to deliver high-quality translations across a wide range of languages and contexts, ultimately improving both the accuracy and readability of the output. Despite its widespread use and significant advancements, as \cite{Karami2014} points out for Google Translate, the LLMs continue to encounter difficulties with more complex texts, idiomatic expressions, and languages that have less digital representation. These limitations underscore the necessity for ongoing improvements and suggest caution when depending on machine translation for critical or nuanced content.

The initial step in addressing the errors identified in the translation of algorithmic and informatics problems from Romanian into English by large language models (LLMs) is to classify them systematically. Classification enables the organization of errors into distinct categories, facilitating a more focused analysis of the translation issues. By grouping errors—such as mistranslations, inconsistencies in technical terminology, or the omission of critical computational terms—it becomes possible to identify recurring patterns and pinpoint areas where LLMs struggle the most. This structured approach lays the foundation for subsequent error quantification and qualitative analysis, essential for improving the overall performance of automated translation systems in technical domains. \cite{Lommel2014} introduce a hierarchical translation error taxonomy known as MQM, which distinguishes between fluency errors and accuracy errors. According to \cite{Lommel2014}, fluency errors are defined as issues "related to the language of the translation, irrespective of its status as a translation," while accuracy errors pertain to "how accurately the target text reflects the content of the source text." Using these categories, errors can be classified as either minor or critical. For this study, errors were categorized based on the classifications established by researchers such as  \cite{Hemchua2007}, \cite{Hsu2014}, \cite{Costa2015}, and according to the frequency of errors identified in the data.

The following table provides a comparative analysis of translation errors identified across various large language models (LLMs) such as Mistral, Llama, Gemma, Gemini, GPT-4o, and Aya. Each model was tasked with translating technical structures from Romanian into English, with the table highlighting the errors, the incorrect target language structure, and the suggested correct translations. While the table does not include every error identified, it highlights those that significantly impact the quality of the translation.

\begin{table}[h!]
    \centering
    \resizebox{\textwidth}{!}{
    \begin{tabular}{|c|c|c|c|}
    \hline
    LLM Model & Problem Number & Structure in Source Language & Error Type \\
    \hline
    Mistral & 515 & rectiliniu & Lexical error \\
    Mistral & 802 & număr de ordine & Semantic error \\
    Llama & 298 & o matrice pătratică & Omission error \\
    Mistral, Llama, Gemma, Gemini, GPT-4o & All texts & „x” & Punctuation error \\
    Mistral, Llama, Gemma, Gemini, GPT-4o & All texts & Tables and their contents & Structural error \\
    GPT-4o & 899 & şir & Semantic error \\
    Aya & 299 & curent & False friend \\
    GPT-4o & 802 & poziţiile de început şi de final ale acestor secvenţe în şirul din set & Semantic error \\
    \hline
    \end{tabular}}
    \caption{Comparative analysis of translation errors across LLM models.}
\end{table}

In this study, several algorithmic problems were translated from Romanian into English by various large language models (LLMs), including Mistral, Llama 3, Llama 3.1, Gemma, Gemini, GPT-4o, and Aya. A consistent pattern of errors and systematic distortions was observed across these translations. Notably, certain parts of the text were left untranslated. For instance, the section "restricții și clarificări" was often only partially translated as "restrictions," omitting "clarifications." Additionally, some tasks were either partially translated or left untranslated entirely, as demonstrated by Llama 3's translation of problem 899. Furthermore, all LLMs failed to fully translate the content of tables, with at least one column from the "example" sections consistently omitted, and the explanatory content entirely absent. In addition to these omissions, punctuation errors were also frequent, with quotation marks, mathematical symbols, and other punctuation marks being incorrectly rendered. These recurring issues highlight the limitations of LLMs in accurately translating structured, technical content.

Several types of translation errors were identified in the output of LLMs when translating algorithmic problems from Romanian into English. One prominent error occurred in the translation of the term "rectiliniu" in problem 515, which was consistently rendered as "straight" by all the LLMs, instead of the correct term "linear". This constitutes a lexical error, as the incorrect word choice alters the intended meaning within the context of the problem. Additionally, this could be classified as a semantic error, given that the LLMs failed to capture the appropriate technical meaning of "rectiliniu" in the domain of algorithmic and computational language, where "linear" is the precise term. The misinterpretation likely arises from the general meaning of "rectiliniu" in everyday language, where "straight" is a common equivalent, but it is not suitable in the technical context, leading to a distortion of the problem's meaning. In the study, Llama exhibited a notable translation error in problem 298 by omitting the adjective "pătratică" when translating the Romanian structure "o matrice pătratică" as "a matrix", rather than the correct "a square matrix". This constitutes a lexical omission error, where a critical modifier—in this case, the adjective "square"—was not translated, resulting in an incomplete and less precise rendering of the source text. The omission impacts the technical accuracy of the translation, as "square matrix" is a specific term in mathematics that conveys essential information about the matrix's dimensions. The failure to include this detail diminishes the clarity and correctness of the translated problem, potentially leading to misinterpretation of the task.

Discrepancies were observed in the translation of key technical terms, particularly the Romanian word "şir", which was inconsistently rendered by different LLMs. For instance, in problem number 899, Llama and GPT-4o translated "şir" as "sequence", while Llama 3 translated it as "string". While both translations capture certain aspects of the term, neither fully aligns with the technical context in which "şir" is used. The most accurate translation in this context would be "array", which is a more precise term in algorithmic and computational problems. According to the literature \cite{Hopcroft2006}, "string" refers to a sequence of characters that can potentially have a very long or infinite length, which may not align with all the contexts given in the algorithmic. The use of "sequence", while closer, still lacks the specificity required to convey the array-like properties of "şir". Thus, the preferred translation of "şir" in such cases would be "array", as it more accurately reflects the intended data structure within the problem's context.

In conclusion, the large language models (LLMs) in this study, especially Llama3.1, Llama 3.2, GPT-4o, and Aya23—showed strong proficiency in translating algorithmic problems from Romanian to English, with overall translation quality approaching that of human translators. However, recurring issues like lexical mismatches, semantic inaccuracies, omissions, and structural inconsistencies highlight the need for human revision to ensure accuracy in technical domains. While LLMs are valuable for initial translations, human oversight remains crucial, especially in specialized contexts like algorithmic problem translation.

\section{Working solution}

We enhanced the OJI Romanian dataset by incorporating manually verified English translations, creating a valuable benchmark for future research in multilingual LLMs. Drawing on extensive experience with Romanian competitive programming, we identified and corrected common machine translation errors, particularly technical jargon often mistranslated by generic tools. Key terms such as "Cerință" (task), "subsecvență" (subarray), "subșir" (subsequence), and "șir de caractere" (string) were carefully handled to ensure accuracy.

Our experiments primarily used GPT-4o and the OpenAI API. We began with a basic prompt—"Can you translate this problem statement into English in the context of competitive programming?"—which, although imperfect, significantly improved focus on competitive programming terminology compared to generic translation approaches.

We further improved the translations by manually correcting recurring errors and refining the prompt based on our expertise with the OJI dataset. This refined prompt ensured a consistent structure in translated problem sets and accommodated Markdown and LaTeX formatting, enhancing clarity and functionality.

For example, a simple prompt that doesn't account for specific terms can lead to pitfalls. Misuse of terms like 'subsequence' and 'substring', or inconsistent descriptions of printed data, can cause confusion. In competitive programming, even one misunderstood word can result in solving an entirely different problem.


\begin{figure}[H]
    \centering
    \includegraphics[width=0.8\linewidth]{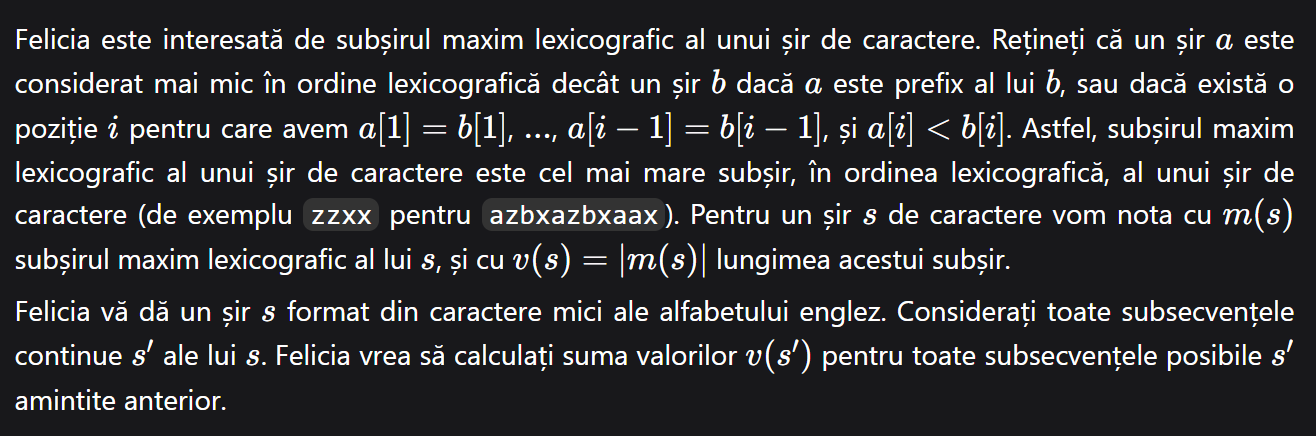}
    \caption{Original Kilonova problem 15}
    \label{fig:subsecvente}
\end{figure}

\begin{figure}[H]
    \centering
    \includegraphics[width=0.8\linewidth]{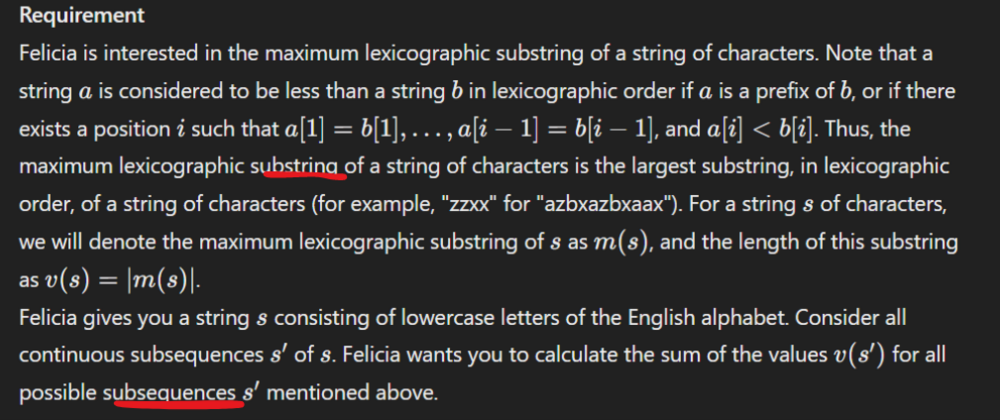}
    \caption{Translated Problem Statement with issues}
    \label{fig:cntsubsirmax}
\end{figure}
 
\subsection{Efficient prompt}

The final version of the prompt used is: \textit{"Please process the following text according to the specified instructions: You will be given a competitive programming problem statement in markdown, written in the Romanian language, using GFM extensions and MathJax/LaTeX math between dollar signs (\$ or \$\$). Another extension is the fact that image attachments are defined using a syntax similar to ~[name.png], with optional attributes named after the end with a vertical bar ('|'). You must translate the statement in the English language, while preserving mathematical values, variable names, general syntax, structure and format. You must also preserve the custom image format exactly as is. The word Cerință is always translated to Task, Date de intrare is translated to Input data, Date de ieșire is translated to Output data, subsecvență is translated to subarray, subșir is translated to subsequence, Restricții și precizări to Constraints and clarifications, vector is translated to array, șir de caractere is translated to string. In addition, in the Date de intrare and Date de ieșire sections, if you see expressions such as Pe prima linie, Pe a doua linie etc., you want to use the verb contain to describe the data we need to read or print. In the Date de ieșire sections, you can also use the verb print to describe the data we need to print. When separating large integers (especially in latex/mathjax math) in groups of 3 digits, do not add a comma. Instead, add a backslash followed and preceded by a single space character. After you are done with translating, please double check the statement and fix potential grammar and/or syntax errors according to the rules of English language."}

Using the refined prompt, the translations became nearly flawless, with no significant errors detected by linguists or members of the competitive programming scientific committee. The translations met both linguistic and technical standards.

The final version of the OJI problem below illustrates the improvements achieved through prompt engineering and domain-specific adjustments while preserving competitive programming terminology.

\begin{figure}[H]
    \centering
    \includegraphics[width=0.8\linewidth]{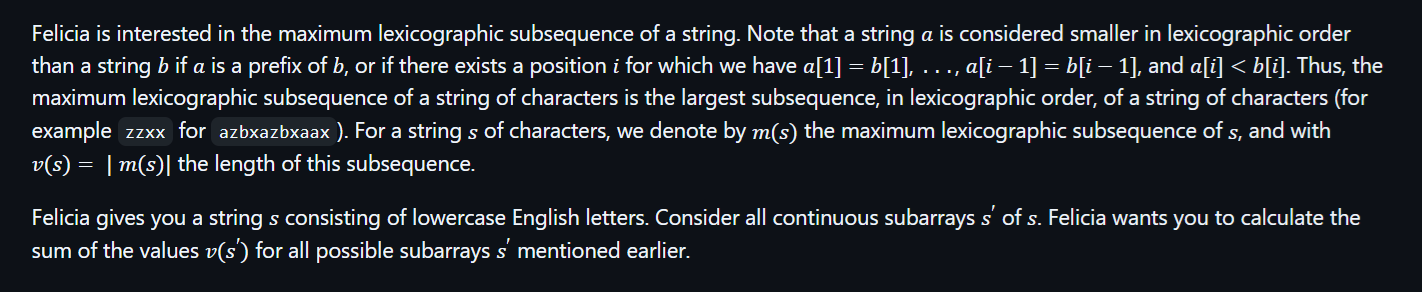}
    \caption{Correct Translation using enhanced prompt.}
    \label{fig:cntsubsirmax2}
\end{figure}

Given the relatively large size of the dataset—over 300 problems, as shown by this list from the \href{https://kilonova.ro/problem_lists/452}{Kilonova online judge}—relying solely on manual browsing was not feasible.

Automating the translation process became essential. We obtained markdown versions of the problem statements from Kilonova, the only Romanian online judge hosting the entire OJI dataset. These files enabled us to develop Python scripts to process the raw statements, interact with the OpenAI API, and improve formatting, reducing potential errors related to the LLM's handling of markdown syntax.

Once the dataset was fully translated, we uploaded it to a \href{https://github.com/stefdasca/statement-translator}{GitHub repository} for evaluating model performance. We further improved the translations by fixing markdown issues and refining specific aspects. This resulted in a comprehensive collection of Romanian problems accurately translated into English, providing a valuable resource for future research and evaluation.

\section{Conclusions}

This study demonstrates that large language models (LLMs) can effectively translate complex technical content from Romanian to English, achieving quality comparable to human translators when provided with well-crafted prompts and human oversight. While focused on Romanian to English, our findings open doors for exploration in other language pairs. Automated translation in STEM competitions is a particularly promising application, where translation quality is crucial and manual translations have historically faced issues of accuracy and potential fraud, highlighting the need for reliable automated solutions.

The results indicate that models like GPT-4.0 and Llama 3.1 8B consistently perform well, though smaller models struggle with complex tasks. Proper prompts and temperature settings (such as 0.6) proved essential for optimal performance. However, despite advancements, lexical errors and structural inconsistencies—especially in smaller models—suggest that human oversight remains necessary.

By enhancing the OJI dataset with manually verified translations, this study provides valuable resources for future research into multilingual LLMs. Expanding these efforts to other language pairs and refining the use of LLMs in high-stakes environments like educational competitions can significantly improve accessibility and fairness in such events.


\end{document}